\documentclass[sigconf]{acmart}

\usepackage[ruled,linesnumbered]{algorithm2e}
\usepackage{makecell}

\AtBeginDocument{%
  }

\setcopyright{acmlicensed}
\copyrightyear{2024}
\acmYear{2024}
\acmDOI{10.1145/3589335.3652503}

\acmConference[WWW '24 Companion]{Companion Proceedings of the ACM Web Conference 2024}{May 13--17, 2024}{Singapore, Singapore.}
\acmBooktitle{Companion Proceedings of the ACM Web Conference 2024 (WWW '24 Companion), May 13--17, 2024, Singapore, Singapore}
\acmISBN{979-8-4007-0172-6/24/05}

\begin{document}

\title{Multi-Granularity Tibetan Textual Adversarial Attack Method Based on Masked Language Model}

\author{Xi Cao}
\orcid{0009-0003-4429-3327}
\affiliation{%
	\institution{School of Information Science and Technology, Tibet University;}
	\institution{Collaborative Innovation Center for Tibet Informatization by MOE and Tibet Autonomous Region}
	\streetaddress{10 Zangda East Road}
	\city{Lhasa}
	\state{Xizang}
	\country{China}
	\postcode{850000}
}
\email{metaphor@outlook.com}

\author{Nuo Qun}
\authornote{Corresponding author.}
\orcid{0000-0003-1984-6770}
\affiliation{%
	\institution{School of Information Science and Technology, Tibet University;}
	\institution{Collaborative Innovation Center for Tibet Informatization by MOE and Tibet Autonomous Region;}
	\institution{Engineering Research Center of Tibetan Information Technology, Ministry of Education}
	\streetaddress{10 Zangda East Road}
	\city{Lhasa}
	\state{Xizang}
	\country{China}
	\postcode{850000}
}
\email{q\_nuo@utibet.edu.cn}

\author{Quzong Gesang}
\orcid{0009-0004-3638-6686}
\affiliation{%
	\institution{School of Information Science and Technology, Tibet University;}
	\institution{Collaborative Innovation Center for Tibet Informatization by MOE and Tibet Autonomous Region}
	\streetaddress{10 Zangda East Road}
	\city{Lhasa}
	\state{Xizang}
	\country{China}
	\postcode{850000}
}
\email{gsqz1212@qq.com}

\author{Yulei Zhu}
\orcid{0009-0001-4812-7196}
\affiliation{%
	\institution{School of Information Science and Technology, Tibet University;}
	\institution{Engineering Research Center of Tibetan Information Technology, Ministry of Education}
	\streetaddress{10 Zangda East Road}
	\city{Lhasa}
	\state{Xizang}
	\country{China}
	\postcode{850000}
}
\email{zhuyulei@utibet.edu.cn}

\author{Trashi Nyima}
\orcid{0009-0001-7393-7449}
\affiliation{%
	\institution{School of Information Science and Technology, Tibet University;}
	\institution{Collaborative Innovation Center for Tibet Informatization by MOE and Tibet Autonomous Region;}
	\institution{Engineering Research Center of Tibetan Information Technology, Ministry of Education}
	\streetaddress{10 Zangda East Road}
	\city{Lhasa}
	\state{Xizang}
	\country{China}
	\postcode{850000}
}
\email{nmzx@utibet.edu.cn}

\renewcommand{\shortauthors}{Xi Cao, Nuo Qun, Quzong Gesang, Yulei Zhu, \&Trashi Nyima}

\begin{abstract}
In social media, neural network models have been applied to hate speech detection, sentiment analysis, etc., but neural network models are susceptible to adversarial attacks.
For instance, in a text classification task, the attacker elaborately introduces perturbations to the original texts that hardly alter the original semantics in order to trick the model into making different predictions.
By studying textual adversarial attack methods, the robustness of language models can be evaluated and then improved.
Currently, most of the research in this field focuses on English, and there is also a certain amount of research on Chinese.
However, there is little research targeting Chinese minority languages.
With the rapid development of artificial intelligence technology and the emergence of Chinese minority language models, textual adversarial attacks become a new challenge for the information processing of Chinese minority languages.
In response to this situation, we propose a multi-granularity Tibetan textual adversarial attack method based on masked language models called {\itshape TSTricker}.
We utilize the masked language models to generate candidate substitution syllables or words, adopt the scoring mechanism to determine the substitution order, and then conduct the attack method on several fine-tuned victim models.
The experimental results show that {\itshape TSTricker} reduces the accuracy of the classification models by more than 28.70\% and makes the classification models change the predictions of more than 90.60\% of the samples, which has an evidently higher attack effect than the baseline method.
\end{abstract}


%


\maketitle

\section{Introduction}

Textual adversarial attack refers to an attack method in which the attacker adds perturbations to the original text that almost do not affect the original semantics, resulting in the incorrect prediction of the language model.
The samples generated by the textual adversarial attack are called adversarial texts.
Studying textual adversarial attack methods can evaluate the robustness of the language model and expose its security vulnerabilities.
Using adversarial texts to train the language model can achieve the effect of improving the model's robustness and actively defending against attacks.

\citet{DBLP:journals/corr/SzegedyZSBEGF13} first found that adding imperceptible noise to the original picture can make the neural network's prediction wrong.
Initially, research on adversarial attacks was limited to the image domain \citep{DBLP:journals/corr/SzegedyZSBEGF13,DBLP:journals/corr/GoodfellowSS14}.
Later, \citet{jia-liang-2017-adversarial} found that language models are also vulnerable to adversarial attacks.
For the reading comprehension system, adding interfering statements to the original paragraph can make the model understand incorrectly.
So far, research on textual adversarial attacks mostly focuses on English \citep{ebrahimi-etal-2018-hotflip,eger-etal-2019-text,DBLP:conf/aaai/JinJZS20,DBLP:conf/aaai/MaheshwaryMP21,DBLP:conf/aaai/YeMWM22,DBLP:conf/aaai/0008X0XZMC0Z23}.
\citet{RJXB201908011} first utilized homophonic word substitution to generate Chinese adversarial texts, and since then some research on textual adversarial attack methods for Chinese has appeared \citep{MESS202302004,DZYX202306035,RJXB202311011}.
Most of the Chinese textual adversarial attacks have considered the characteristics of the Chinese language itself.
For example, a simplified Chinese character can be converted into a traditional Chinese character, rewritten into Chinese Pinyin, or split into Chinese character components.
Moreover, native Chinese speakers have a certain degree of tolerance towards Chinese characters that are similar in pronunciation or shape.
However, there is currently little relevant research on Chinese minority languages.

With the development of artificial intelligence technology and the construction of informatization in Tibetan areas of China, PLMs (pre-trained language models) targeting or including Tibetan have appeared in the past two years, such as monolingual models Tibetan-BERT \citep{DBLP:conf/iccir/ZhangKGTQ22} and TiBERT \citep{9945074}, and multilingual models CINO \citep{yang-etal-2022-cino}, MiLMo \citep{10393961} and CMPT \citep{DBLP:conf/ccmt/LiWSL22}, etc.
Tibetan language models obtained through the paradigm of ``{\itshape PLMs + Fine-tuning}'' can achieve good results in downstream tasks.
Recently, \citet{cao-etal-2023-pay-attention} proposed the first textual adversarial attack method targeting Tibetan and called attention to the robustness and security of Chinese minority language models.
They used cosine similarity to generate candidate substitution syllables and implemented their attack method on one kind of language model.
Textual adversarial attack is a new challenge for the information processing of Chinese minority languages.
The study of textual adversarial attack methods targeting the Tibetan language is of practical significance for the informatization construction and stable development of Tibetan areas in China.

The main contributions of this paper are as follows:

(1) This paper proposes a multi-granularity Tibetan textual adversarial attack method based on masked language models called {\itshape TSTricker} which can attack from syllable and word granularity.
The attack effect is significantly stronger than the baseline method.

(2) This paper implements the attack method on the BERT-based monolingual model targeting Tibetan and the XLMRoBERTa-based multilingual model including Tibetan.
The experimental results show the vulnerability of these language models in terms of robustness.

(3) This paper finds that  using publicly available PLMs can effectively carry out textual adversarial attacks on language models in downstream tasks, providing theoretical guidance for active defense in the future.

(4) This paper calls more attention to the security issues in the information processing of Chinese minority languages. In order to facilitate further research, we make the models involved in this paper publicly available on Hugging Face (\url{https://huggingface.co/UTibetNLP}) and the code involved in this paper publicly available on GitHub (\url{https://github.com/UTibetNLP/TSTricker}).

\section{Preliminaries}

\subsection{Textual Adversarial Attacks on Text Classification}

For a text classification task, let $x$ be the original input text ($x\in{X}$, $X$ includes all possible input texts), $F$ be the classifier, and the output label of $F$ on $x$ be $y$ ($y\in{Y}$, $Y$ includes all possible output labels), denoted as
\begin{equation}
	\label{equation1}
	F(x)={arg\max_{y^*\in{Y}}{P(y^*|x)}}={y}.
\end{equation}
Let $\delta$ be the perturbation on $x$ and $x'$ be the input text after the perturbation, then $x' =x+\delta$.
By elaborately designing $\delta$, the attacker makes the semantics of the input text basically unchanged.
The attacker achieves a successful textual adversarial attack when
\begin{equation}
	\label{equation2}
	F(x')={arg\max_{y^*\in{Y}}{P(y^*|x')}}\neq{y}.
\end{equation}

\subsection{Characteristics of Tibetan Script}
\label{subsection2.2}

\begin{figure}[h]
	\centering
	\includegraphics[width=\linewidth]{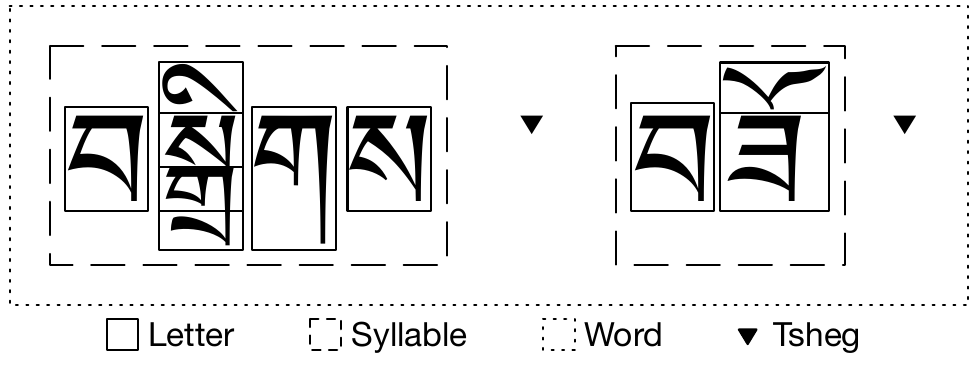}
	\caption{Structure of Tibetan Word {\itshape Programming}}
	\label{figure1}
\end{figure}

Tibetan is a phonetic script composed of 30 consonant letters and 4 vowel letters, each of which has its corresponding Unicode.
These letters are spelled into Tibetan syllables by strict rules, and one or more Tibetan syllables form Tibetan words, separated by Tibetan separators called Tsheg.
Therefore, the granularity of Tibetan script, from small to large, is divided into letters [or Unicodes], syllables, and words, which is different from English (letters [or Unicodes], words) and Chinese (characters [or Unicodes], words).
Figure \ref{figure1} illustrates the structure of a Tibetan word clearly.
Let the syllable in the original input text $x$ be $s$ (ignore Tsheg) and the word be $w$, then
\begin{equation}
	\label{equation3}
	x=s_{1}s_{2}\dots{s_{i}}\dots{s_{m}},
\end{equation}
\begin{equation}
	\label{equation4}
	x=w_{1}w_{2}\dots{w_{j}}\dots{w_{n}}.
\end{equation}

\section{Attack Method}

In this section, we will detail the {\itshape TSTricker} algorithm, a multi-granularity Tibetan textual adversarial attack method based on masked language models. 
We utilize the masked language models to generate candidate substitution syllables or words and adopt the scoring mechanism to determine substitution order.

\subsection{Generate Substitutions}

Tibetan-BERT\footnote{\url{https://huggingface.co/UTibetNLP/tibetan_bert}} \citep{DBLP:conf/iccir/ZhangKGTQ22} and TiBERT\footnote{\url{http://tibert.cmli-nlp.com/}} \citep{9945074} are both BERT-based Tibetan monolingual PLMs, but their tokenizers are different.
The tokenizer granularity of Tibetan-BERT is syllable and sub-syllable, while the tokenizer granularity of TiBERT is word, sub-word, and syllable (including Tsheg).
Tibetan-BERT and TiBERT as masked language models can well predict the missing syllables and words of the text, respectively, and make the semantics unobstructed.
In this work, we use Tsheg and the word segmentation tool TibetSegEYE\footnote{\url{https://github.com/yjspho/TibetSegEYE}} to segment the original input text into syllables and words.

\begin{figure*}[h]
	\centering
	\includegraphics[width=\linewidth]{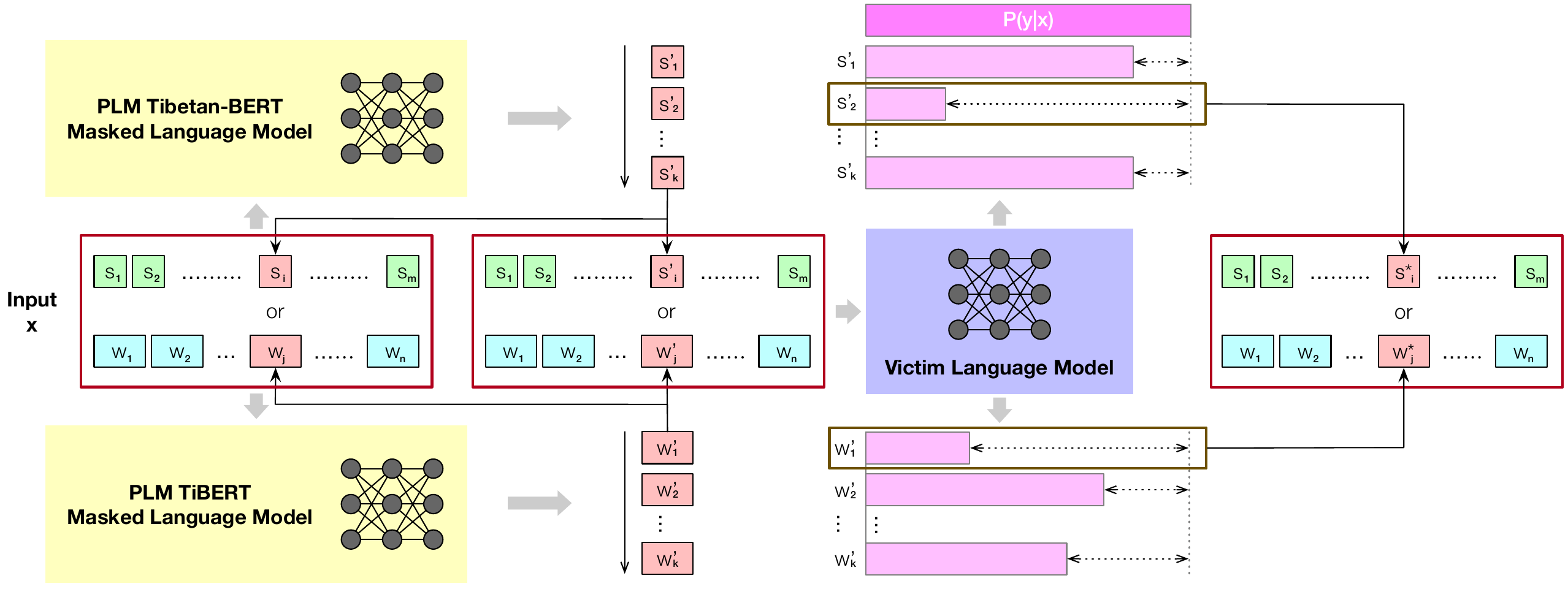}
	\caption{Process of Generating Substitutions}
	\label{figure2}
\end{figure*}

For each syllable $s$ and word $w$ in the original input text $x$, we use Tibetan-BERT and TiBERT to predict the possible top-$k$ syllables and words other than the original syllable and word as the set of candidate substitution syllables $C_s$ and the set of candidate substitution words $C_w$, respectively.
In our experiment, $k$ is set to 50 by default, and syllables are filtered out for the prediction results of Tibetan-BERT and words are filtered out for the prediction results of TiBERT.
After that, we iterate over the syllable $s'$ in $C_s$ or the word $w'$ in $C_w$, then
\begin{equation}
	\label{equation5}
	x'=s_{1}s_{2}\dots{{s_{i}}'}\dots{s_{m}}~or~x'=w_{1}w_{2}\dots{{w_{j}}'}\dots{w_{n}},
\end{equation}
and calculate
\begin{equation}
	\label{equation6}
	\Delta{P}=P(y|x)-P(y|x').
\end{equation}
After the iteration is completed, the syllable $s^*$ or the word $w^*$ can be always found, and
\begin{equation}
	\label{equation7}
	x^*=s_{1}s_{2}\dots{{s_{i}}^*}\dots{s_{m}}~or~x^*=w_{1}w_{2}\dots{{w_{j}}^*}\dots{w_{n}},
\end{equation}
at which point,
\begin{align}
	\label{equation8}
	\Delta{P^*}&=P(y|x)-P(y|{x^*}) \\
	&=max\{\Delta{P_k\}}_{k=1}^{size(C_s~or~C_w)} \notag \\
	&=max\{P(y|x)-P(y|{{x'}_k})\}_{k=1}^{size(C_s~or~C_w)}, \notag
\end{align}
\begin{align}
	\label{equation9}
	{s^*}~or~{w^*}&=arg\max_{{s'}\in{C_{s}}\atop{{w'}\in{C_{w}}}}\{\Delta{P_k\}}_{k=1}^{size(C_s~or~C_w)} \\
	&=arg\max_{{s'}\in{C_{s}}\atop{{w'}\in{C_{w}}}}\{P(y|x)-P(y|{x'}_{k})\}_{k=1}^{size(C_s~or~C_w)}. \notag
\end{align}
It means that substituting the syllable $s$ with $s^*$ or the word $w$ with $w^*$ can result in the largest decrease in classification probability and the best attack effect.
Figure \ref{figure2} illustrates this process.

\subsection{Determine Substitution Order}

\begin{figure}[h]
	\centering
	\includegraphics[width=\linewidth]{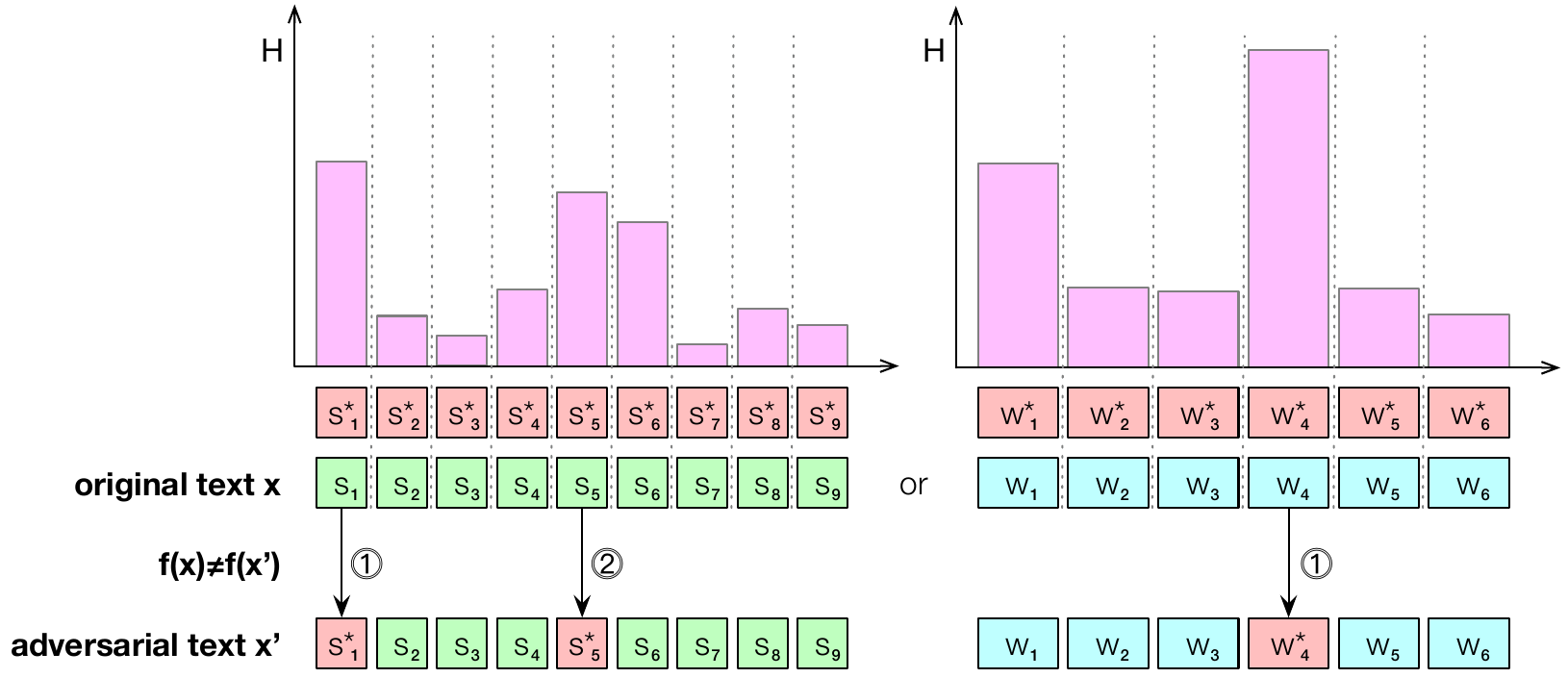}
	\caption{Process of Determining Substitution Order}
	\label{figure3}
\end{figure}

\citet{DBLP:journals/corr/LiMJ16a} proposed {\itshape word saliency}, which refers to the drop value of the classification probability after setting a word in the original input text to unknown.
We use this metric to calculate the saliency of a syllable or word.
Let
\begin{equation}
	\label{equation10}
	\hat{x}=s_{1}s_{2}\dots{unk.}\dots{s_{m}}~or~\hat{x}=w_{1}w_{2}\dots{unk.}\dots{w_{n}},
\end{equation}
\begin{equation}
	\label{equation11}
	S=P(y|x)-P(y|\hat{x}).
\end{equation}
\citet{ren-etal-2019-generating} further proposed {\itshape probability weighted word saliency} and the metric is as follows:
\begin{equation}
	\label{equation12}
	H=Softmax(S)\cdot\Delta{P^{*}},
\end{equation}
which takes into account the saliency of both the substituted word and the substitution word.
We use this metric to decide the order of substituting syllables or words.
After sorting all the scores $\{H_{1},H_{2},\dots\}$ corresponding to the original input text $x$ in descending order, we sequentially substitute the syllable $s_i$ with ${s_i}^*$ or substitute the word $w_j$ with ${w_j}^*$ until the output label changes, then the attack succeeds.
Otherwise, the attack fails.
Figure \ref{figure3} illustrates this process.

\subsection{Pseudocode}
The pseudocode of the syllable-level {\itshape TSTricker} algorithm is as shown in Algorithm \ref{algorithm1}.

\begin{algorithm}[h]
	\caption{Syllable-Level TSTricker Algorithm}
	\label{algorithm1}
	\KwIn{Classifier $F$.}
	\KwIn{Original text $x=s_{1}s_{2}\dots{s_{i}}\dots{s_{m}}$.}
	\KwOut{Adversarial text $x'$.}
	\For{$i\leftarrow1$ to $m$}
	{
		// Prepare to calculate $Softmax(S_{i})$ in line 15. \\
		$\hat{{x}_{i}}\leftarrow{s_{1}s_{2}\dots{unk.}\dots{s_{m}}}$\tcp*[f]{Equation \ref{equation10}} \\
		$S_{i}\leftarrow{P(y|x)-P(y|\hat{x_{i}})}$\tcp*[f]{Equation \ref{equation11}}
	}
	Init $L$ as an empty list. \\
	\For{$i\leftarrow1$ to $m$}
	{
		Get ${C_{s}}_{i}$, the set of candidate substitution syllables, according to Tibetan-BERT. \\
		\ForEach{${s_{i}}'$ in ${C_{s}}_{i}$}
		{
			{${x_{i}}'\leftarrow{s_{1}s_{2}\dots{{s_{i}}'}\dots{s_{m}}}$}\tcp*[f]{Equation \ref{equation5}} \\
			$\Delta{P_{i}}\leftarrow{P(y|x)-P(y|{x_{i}}')}$\tcp*[f]{Equation \ref{equation6}}
		}
		$\Delta{{P_{i}}^{*}}\leftarrow{max\{\Delta{P_{ik}\}}_{k=1}^{len({C_{s}}_{i})}}$\tcp*[f]{Equation \ref{equation8}} \\
		${s_{i}}^{*}\leftarrow{arg\max_{{{s_{i}}'}\in{{C_{s}}_{i}}}\{\Delta{P_{ik}\}}_{k=1}^{len({C_{s}}_{i})}}$\tcp*[f]{Equation \ref{equation9}} \\
		$H_{i}\leftarrow{Softmax(S_{i})\cdot\Delta{{P_{i}}^{*}}}$\tcp*[f]{Equation \ref{equation12}} \\
		Append $({s_{i}}^{*},H_{i})$ into $L$.
	}
	Sort $L$ by the second parameter in descending order. \\
	\ForEach{$({s_{i}}^{*},H_{i})$ in $L$}
	{
		$x'\leftarrow{s_{1}s_{2}\dots{{s_{i}}^{*}}\dots{s_{m}}}$ \\
		\If{${F(x')}\neq{F(x)}$}
		{
			Attack succeeds and return $x'$.
		}
	}
	Attack fails and return.
\end{algorithm}

\section{Experiment}

\subsection{Victim Language Models}

\begin{figure}[h]
	\centering
	\includegraphics[width=\linewidth]{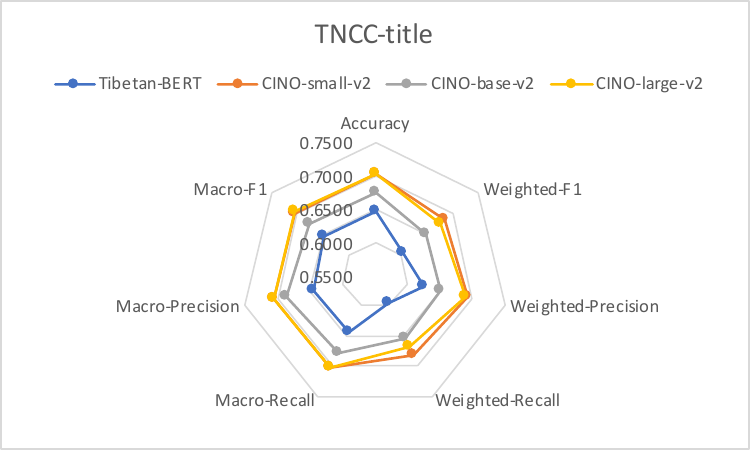}
	\includegraphics[width=\linewidth]{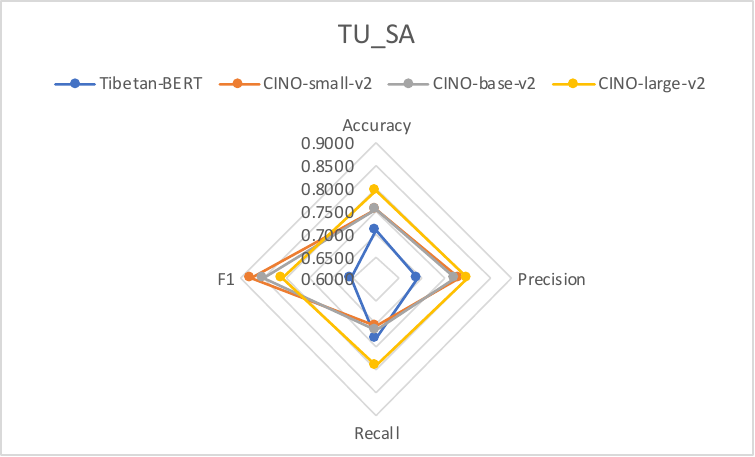}
	\caption{Performance of Victim Language Models}
	\label{figure4}
\end{figure}

We use the paradigm of ``{\itshape PLMs + Fine-tuning}'' to construct victim language models.
We divide each dataset in the downstream tasks into a training set, a validation set, and a test set according to the ratio of 8:1:1.
On the training set, we fine-tune the PLMs to generate the victim language models, and on the test set, we conduct the attack method on the victim language models.
We select the Tibetan news title classification dataset TNCC-title \citep{DBLP:conf/cncl/QunLQH17} and the Tibetan sentiment analysis dataset TU\_SA \citep{MESS202302007}.
We select Tibetan-BERT \citep{DBLP:conf/iccir/ZhangKGTQ22}, a monolingual PLM targeting Tibetan, and CINO series \citep{yang-etal-2022-cino}, a series of multilingual PLM including Tibetan.
At last, 8 victim language models are constructed which we make them publicly available on Hugging Face (\url{https://huggingface.co/UTibetNLP}) for future research.
Table \ref{table1} lists the detailed information of the datasets.
Table \ref{table2} lists the hyperparameters of the fine-tuning.
Table \ref{table3} and Table \ref{table4} respectively list the performance of the models on TNCC-title and TU\_SA.
Figure \ref{figure4} also presents the performance of these victim language models in the form of radar charts.
The following is a brief introduction to the datasets and the PLMs.

\subsubsection{Datasets}

\begin{itemize}
	\item \textbf{TNCC-title\footnote{\url{https://github.com/FudanNLP/Tibetan-Classification}}.}
	A Tibetan news title classification dataset. \citet{DBLP:conf/cncl/QunLQH17} collect the dataset from the China Tibet Online website (\url{http: //tb.tibet.cn}).
	This dataset contains a total of 9,276 Tibetan news titles, which are divided into 12 classes: politics, economics, education, tourism, environment, language, literature, religion, arts, medicine, customs, and instruments.
	\item \textbf{TU\_SA\footnote{\url{https://github.com/UTibetNLP/TU_SA}}.}
	A Tibetan sentiment analysis dataset. \citet{MESS202302007} build the dataset by translating and proofreading 10,000 sentences from two public simplified Chinese sentiment analysis datasets  weibo\_senti\_100k and ChnSentiCorp (\url{https://github.com/SophonPlus/ChineseNlpCorpus}).
	In this dataset, negative and positive classes each account for 50\%.
\end{itemize}

\begin{table*}
	\centering
	\caption{Information of Datasets}
	\label{table1}
	\begin{tabular}{ccccccccc}
		\toprule
		Dataset & Task & \#Classes & \makecell{\#Average\\Syllables} & \makecell{\#Average\\Letters} & \makecell{\#Total\\Samples} & \makecell{\#Training\\Samples} & \makecell{\#Validation\\Samples} & \makecell{\#Test\\Samples} \\
		\midrule
		TNCC-title & \makecell{news title\\classification} & 12 & 15.5031 & 63.6196 & 9,276 & 7,422 & 927 & 927 \\
		\hline
		TU\_SA & \makecell{sentiment\\analysis} & 2 & 27.8724 & 108.1897 & 10,000 & 8,000 & 1,000 & 1,000 \\
		\bottomrule
	\end{tabular}
\end{table*}

\subsubsection{PLMs}

\begin{itemize}
	\item \textbf{Tibetan-BERT\footnote{\url{https://huggingface.co/UTibetNLP/tibetan_bert}}.}
	A BERT-based monolingual PLM targeting Tibetan.
	\citet{DBLP:conf/iccir/ZhangKGTQ22} propose and release the first Tibetan BERT PLM, which achieves a good result on the specific downstream Tibetan text classification dataset.
	\item \textbf{CINO\footnote{\url{https://huggingface.co/hfl/cino-small-v2}}\footnote{\url{https://huggingface.co/hfl/cino-base-v2}}\footnote{\url{https://huggingface.co/hfl/cino-large-v2}}.}
	A XLMRoBERTa-based multilingual PLM including Tibetan.
	\citet{yang-etal-2022-cino} propose and release the first multilingual PLM for Chinese minority languages, covering Standard Chinese, Yue Chinese and six ethnic minority languages.
	It achieves SOTA performance on multiple monolingual or multilingual text classification datasets.
\end{itemize}

\begin{table*}
	\centering
	\caption{Hyperparameters of Fine-tuning}
	\label{table2}
	\begin{tabular}{ccccccc}
		\toprule
		Model & Dataset & Batch Size & Epochs & Learning Rate & Warmup Ratio & Metric for Best Model \\
		\midrule
		Tibetan-BERT & TNCC-title \& TU\_SA & 32 & 20 & 5e-5 & 0.0 & Macro-F1 \& F1 \\
		CINO-small-v2 & TNCC-title \& TU\_SA & 32 & 40 & 5e-5 & 0.1 & Macro-F1 \& F1 \\
		CINO-base-v2 & TNCC-title \& TU\_SA & 32 & 40 & 5e-5 & 0.1 & Macro-F1 \& F1 \\
		CINO-large-v2 & TNCC-title \& TU\_SA & 32 & 40 & 3e-5 & 0.1 & Macro-F1 \& F1 \\
		\bottomrule
	\end{tabular}
\end{table*}

\begin{table*}
	\centering
	\caption{Model Performance on TNCC-title}
	\label{table3}
	\begin{tabular}{cccccccc}
		\toprule
		Model & Accuracy & \makecell{Macro-\\F1} & \makecell{Macro-\\Precision} & \makecell{Macro-\\Recall} & \makecell{Weighted-\\F1} & \makecell{Weighted-\\Precision} & \makecell{Weighted-\\Recall} \\
		\midrule
		Tibetan-BERT & 0.6462 & 0.6057 & 0.6251 & 0.5956 & 0.6423 & 0.6450 & 0.6462 \\
		CINO-small-v2 & \textbf{0.7023} & \underline{\textbf{0.6839}} & \underline{\textbf{0.6918}} & \underline{\textbf{0.6819}} & \textbf{0.7016} & \underline{\textbf{0.7069}} & \textbf{0.7023} \\
		CINO-base-v2 & 0.6764 & 0.6488 & 0.6523 & 0.6556 & 0.6772 & 0.6853 & 0.6764 \\
		CINO-large-v2 & \underline{\textbf{0.7044}} & \textbf{0.6759} & \textbf{0.6898} & \textbf{0.6672} & \underline{\textbf{0.7025}} & \textbf{0.7062} & \underline{\textbf{0.7044}} \\
		\bottomrule
	\end{tabular}
	\\\underline{\textbf{Bold and underlined values}} represent \underline{\textbf{the best performance}}.
	\\\textbf{Bold values} represent \textbf{the second best performance}.
\end{table*}

\begin{table}
	\centering
	\caption{Model Performance on TU\_SA}
	\label{table4}
	\begin{tabular}{ccccc}
		\toprule
		Model & Accuracy & F1 & Precision & Recall  \\
		\midrule
		Tibetan-BERT & 0.7070 & 0.6913 & \textbf{0.7305} & 0.6560 \\
		CINO-small-v2 & \textbf{0.7550} & \textbf{0.7818} & 0.7047 & \underline{\textbf{0.8780}} \\
		CINO-base-v2 & 0.7530 & 0.7748 & 0.7119 & \textbf{0.8500} \\
		CINO-large-v2 & \underline{\textbf{0.7970}} & \underline{\textbf{0.7992}} & \underline{\textbf{0.7906}} & 0.8080 \\
		\bottomrule
	\end{tabular}
	\\\underline{\textbf{Bold and underlined values}} represent \underline{\textbf{the best performance}}.
	\\\textbf{Bold values} represent \textbf{the second best performance}.
\end{table}

\subsection{Evaluation Metrics}

\begin{itemize}
	\item \textbf{Accuracy Drop Value ({\itshape ADV}).}
	$ADV$ refers to the drop value of the model accuracy post-attack compared to pre-attack, as defined in Equation \ref{equation13}, with which we evaluate the attack effectiveness and the model robustness.
	The larger $ADV$, the more effective the attack, and the less robust the model.
	\begin{equation}
		\label{equation13}
		ADV=Accuracy_{pre}-Accuracy_{post}
	\end{equation}
	\item \textbf{Attack Success Ratio ({\itshape ASR}).}
	$ASR$ refers to the ratio of the number of changes in model predictions before and after the attack, as defined in Equation \ref{equation14}, which we use to evaluate the attack effectiveness and the model robustness.
	The larger $ASR$, the more effective the attack, and the less robust the model.
	\begin{equation}
		\label{equation14}
		ASR=\frac{Num_{F(x)\neq{F(x')}}}{Num_{total}}
	\end{equation}
	\item \textbf{Levenshtein Distance (\itshape LD).}
	$LD$ between the original text and the adversarial text is the minimum number of single-letter edits (insertions, deletions, or substitutions) required to change one text into the other, as defined in Equation \ref{equation15}.
	We use this metric to evaluate the quality of the adversarial text, and the smaller $LD$, the better the quality of the adversarial text.
	\begin{equation}
		\label{equation15}
		LD_{x,x'}(i,j)=\begin{cases}
			\max(i,j) & if~min(i,j)=0,\\
			LD_{x,x'}(i-1,j-1) & if~x_i=x'_j,\\
			1+\min\begin{cases}
				LD_{x,x'}(i-1,j)\\
				LD_{x,x'}(i,j-1)\\
				LD_{x,x'}(i-1,j-1)
			\end{cases} & otherwise.
		\end{cases}
	\end{equation}
\end{itemize}

\subsection{Experiment Results}

\begin{figure}[h]
	\centering
	\includegraphics[width=\linewidth]{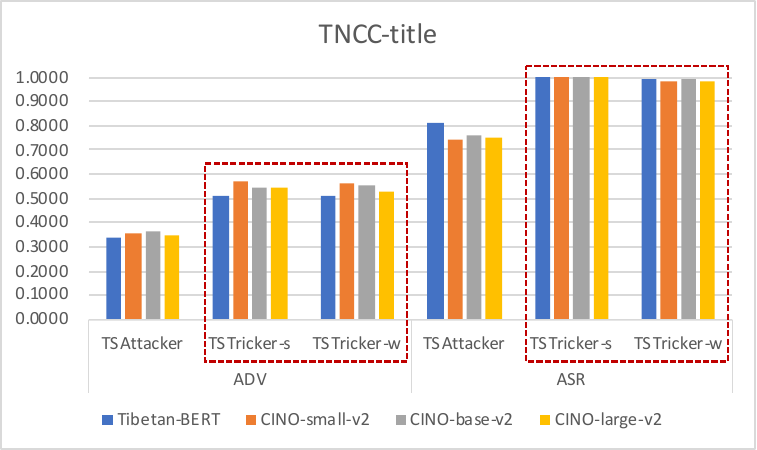}
	\includegraphics[width=\linewidth]{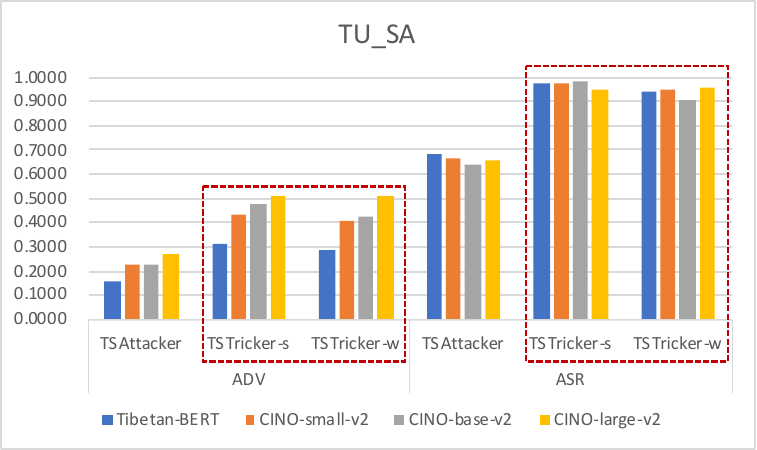}
	\includegraphics[width=\linewidth]{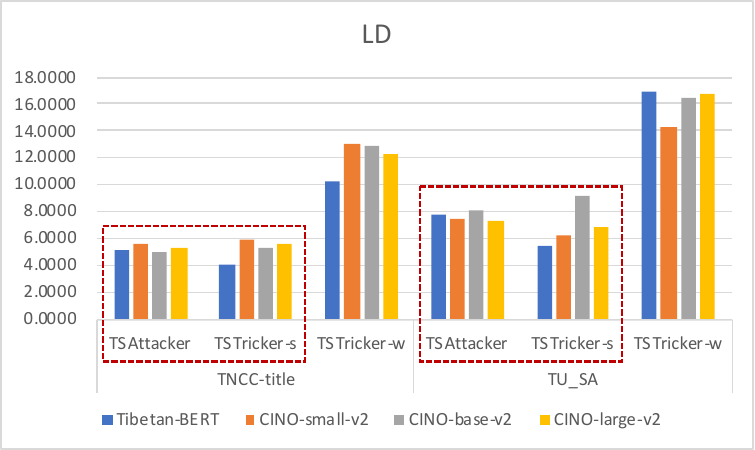}
	\caption{Performance of Attacks}
	\label{figure5}
\end{figure}

\citet{cao-etal-2023-pay-attention} propose the first Tibetan textual adversarial attack method called {\itshape TSAttacker}, which we use as a baseline method for comparison.
Table \ref{table5} and Table \ref{table6} list the performance of involved attack methods against the victim language models on the TNCC-title and TU\_SA datasets, respectively.
{\itshape TSTricker-s} represents syllable-level {\itshape TSTricker}, and {\itshape TSTricker-w} represents word-level {\itshape TSTricker}.
Figure \ref{figure5} clearly depicts the attack performance.
Table \ref{table7} lists several generated adversarial text.

From the perspective of $ADV$ and $ASR$, {\itshape TSTricker} has a significant improvement in attack effect compared with {\itshape TSAttacker}, and both syllable-level and word-level {\itshape TSTricker} can make the model change the predictions of more than 90\% of the samples.
From the perspective of $LD$, {\itshape TSAttacker} and {\itshape TSTricker-s} are syllable-level attacks, while {\itshape TSTricker-w} is a word-level attack, so a larger attack granularity leads to a larger LD.
As introduced in Subsection \ref{subsection2.2}, the granularity of Tibetan script, from small to large, is divided into letters [or Unicodes], syllables, and words, therefore the granularity of word-level attacks is larger than syllable-level attacks.
In Table \ref{table7}, on TNCC-title, {\itshape TSTricker-s} and {\itshape TSTricker-w} both change the Tibetan word ``traveling'', and on TU\_SA, {\itshape TSTricker-s} and {\itshape TSTricker-w} both change the Tibetan word ``cute''.
Native Tibetan speakers find the semantics of the {\itshape TSTricker-s} generated adversarial texts not changed, while the semantics of the {\itshape TSTricker-w} generated adversarial texts changed.
Also, word-level {\itshape TSTricker} depends on the effectiveness of the segmentation tool.
In summary, {\itshape TSTricker-s} is a good attack method with a high attack effect and fewer perturbations, while {\itshape TSTricker-w} has larger perturbation granularity and may cause semantics change.

\begin{table*}
	\centering
	\caption{Attack Performance on TNCC-title}
	\label{table5}
	\begin{tabular}{cccccc}
		\toprule
		Metric & Attack & Tibetan-BERT & CINO-small-v2 & CINO-base-v2  & CINO-large-v2 \\
		\midrule
		~ & TSAttacker & 0.3420 & 0.3592 & 0.3646 & 0.3430 \\
		\multirowcell{3}{$ADV$} & TSTricker-s & \underline{\textbf{0.5124}} & \underline{\textbf{0.5685}} & \textbf{0.5414} & \underline{\textbf{0.5426 }}\\
		~ & TSTricker-w & \underline{\textbf{0.5124}} & \textbf{0.5588} & \underline{\textbf{0.5566}} & \textbf{0.5286} \\
		\midrule
		~ & TSAttacker & 0.8090 & 0.7400 & 0.7605 & 0.7487 \\
		\multirowcell{3}{$ASR$} & TSTricker-s & \underline{\textbf{0.9968}} & \underline{\textbf{0.9989}} & \underline{\textbf{0.9989}} & \underline{\textbf{0.9978}} \\
		~ & TSTricker-w & \textbf{0.9957} & \textbf{0.9806} & \textbf{0.9935} & \textbf{0.9817} \\
		\midrule
		~ & TSAttacker & \textbf{5.2000} & \underline{\textbf{5.6210}} & \underline{\textbf{5.0638}} & \underline{\textbf{5.3386}} \\
		\multirowcell{3}{$LD$} & TSTricker-s & \underline{\textbf{4.0671}} & \textbf{5.8856} & \textbf{5.3402} & \textbf{5.6865} \\
		~ & TSTricker-w & 10.2492 & 13.0297 & 12.9511 & 12.3374 \\
		\bottomrule
	\end{tabular}
	\\\underline{\textbf{Bold and underlined values}} represent \underline{\textbf{the best performance}}.
	\\\textbf{Bold values} represent \textbf{the second best performance}.
\end{table*}

\begin{table*}
	\centering
	\caption{Attack Performance on TU\_SA}
	\label{table6}
	\begin{tabular}{cccccc}
		\toprule
		Metric & Attack & Tibetan-BERT & CINO-small-v2 & CINO-base-v2  & CINO-large-v2 \\
		\midrule
		~ & TSAttacker & 0.1570 & 0.2260 & 0.2240 & 0.2660 \\
		\multirowcell{3}{$ADV$} & TSTricker-s & \underline{\textbf{0.3080}} & \underline{\textbf{0.4300}} & \underline{\textbf{0.4730}} & \textbf{0.5060} \\
		~ & TSTricker-w & \textbf{0.2870} & \textbf{0.4050} & \textbf{0.4200} & \underline{\textbf{0.5100}} \\
		\midrule
		~ & TSAttacker & 0.6810 & 0.6640 & 0.6380 & 0.6570 \\
		\multirowcell{3}{$ASR$} & TSTricker-s & \underline{\textbf{0.9740}} & \underline{\textbf{0.9780}} & \underline{\textbf{0.9870}} & \textbf{0.9520} \\
		~ & TSTricker-w & \textbf{0.9390} & \textbf{0.9470} & \textbf{0.9060} & \underline{\textbf{0.9560}} \\
		\midrule
		~ & TSAttacker & \textbf{7.7298} & \textbf{7.4533} & \underline{\textbf{8.0769}} & \textbf{7.3369} \\
		\multirowcell{3}{$LD$} & TSTricker-s & \underline{\textbf{5.4887}} & \underline{\textbf{6.2495}} & \textbf{9.2057} & \underline{\textbf{6.8813}} \\
		~ & TSTricker-w & 16.9542 & 14.2365 & 16.5066 & 16.7699 \\
		\bottomrule
	\end{tabular}
	\\\underline{\textbf{Bold and underlined values}} represent \underline{\textbf{the best performance}}.
	\\\textbf{Bold values} represent \textbf{the second best performance}.
\end{table*}

\begin{table*}
	\centering
	\caption{Several Generated Adversarial Texts}
	\label{table7}
	\begin{tabular}{ccc}
		\toprule
		Dataset & Text & Class \\
		\midrule
		~ & \makecell{Original Text:\\\includegraphics[scale=0.5]{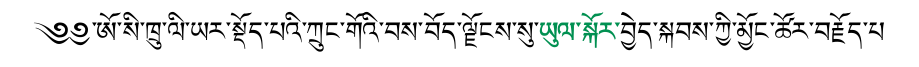}\\English Translation:\\Chinese in Australia talking about their feelings when \textbf{traveling} in Xizang.} & \makecell{Tourism\\(99.5900\%)} \\
		\cline{2-3}
		\multirowcell{3}{TNCC-title} & \makecell{{\itshape TSTricker-s} Generated Adversarial Text:\\\includegraphics[scale=0.5]{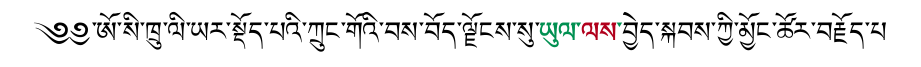}}  & \makecell{Politics\\(55.0304\%)} \\
		\cline{2-3}
		~  & \makecell{{\itshape TSTricker-w} Generated Adversarial Text:\\\includegraphics[scale=0.5]{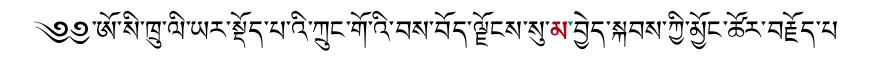}} & \makecell{Politics\\(94.4312\%)} \\
		\midrule
		~ & \makecell{Original Text:\\\includegraphics[scale=0.5]{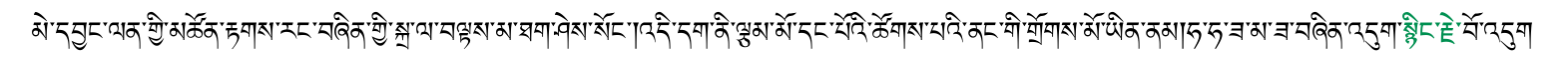}\\English Translation:\\Yanglan Mei's iconic hair is evident at a glance. Are these the best friends in the First Lady circle?\\Haha, everyone is eating, it's incredibly \textbf{cute}.} & \makecell{Positive\\(99.9726\%)} \\
		\cline{2-3}
		\multirowcell{3}{TU\_SA} & \makecell{{\itshape TSTricker-s} Generated Adversarial Text:\\\includegraphics[scale=0.5]{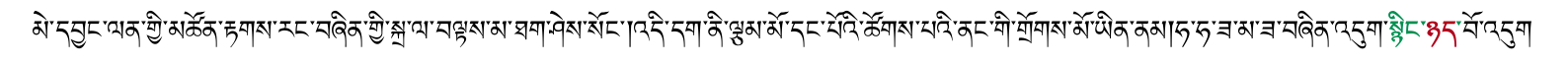}}  & \makecell{Negative\\(99.6808\%)} \\
		\cline{2-3}
		~  & \makecell{{\itshape TSTricker-w} Generated Adversarial Text:\\\includegraphics[scale=0.5]{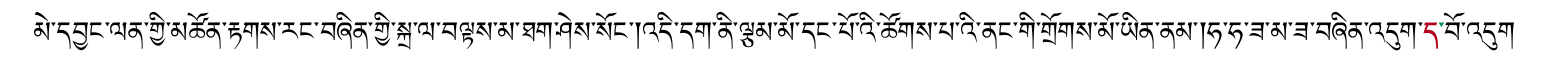}} & \makecell{Negative\\(99.8371\%)} \\
		\bottomrule
	\end{tabular}
\end{table*}

\section{Discussion}

\subsection{Adversarial Attack in Social Media}

In social media, neural network models have been used for hate speech detection, pornographic content detection, etc., and have achieved good results.
However, adversarial content is still ubiquitous on the internet, and adversarial content can always evade detection by various means.
Studying adversarial attacks in social media is crucial to purify the network.
\citet{DBLP:journals/corr/abs-2302-12095} evaluate the adversarial robustness of ChatGPT and find that even for ChatGPT, its absolute performance is far from perfection.
Therefore, there is still enormous exploration room for research on the robustness and interpretability of neural network models.

\subsection{Robustness of Chinese Minority Language Models}

Due to the lag of Chinese minority natural language processing technology, the conditions for studying the robustness of Chinese minority language models have just matured.
Textual adversarial attack is a new challenge.
China is a unified multi-ethnic country, and studying textual adversarial attacks plays an important role in the informatization construction and stable development of ethnic minority areas in China.

\section{Conclusion and Future Work}

This paper proposes a multi-granularity Tibetan textual adversarial attack method based on masked language models called {\itshape TSTricker}, which can reduce the accuracy of the victim models by more than 28.70\% and make the victim models change the predictions of more than 90.60\% of the samples.
This attack method has an evidently higher attack effect than the baseline method.
In particular, syllable-level {\itshape TSTricker} performs excellently in attack effectiveness, perturbation ratio, and semantics preservation.
However, word-level {\itshape TSTricker} may not perform well at the level of perturbation ratio and semantics preservation.
It does not mean that adversarial attacks closer to real attack scenarios are more useful, as white-box adversarial attacks and soft-label black-box adversarial attacks have been used to reinforce models before they go live.
Accordingly, we believe that {\itshape TSTricker-s} can be well applied to Tibetan NLP models' reinforcement scenarios.

We will divide future work into two parts.
Part of it is active defense, and we will explore more Tibetan textual adversarial attack methods to generate a Tibetan adversarial dataset and construct a Tibetan textual adversarial attack toolkit.
On the one hand, the dataset can be used to evaluate the robustness of Tibetan language models, and on the other hand, the toolkit can be used for adversarial training to improve the robustness of Tibetan language models.
The other part is to explore hard-label or more efficient textual adversarial attack methods, providing theoretical guidance for active defense.

\section{Ethics Statement}

This paper aims to demonstrate the effectiveness of our attack method and the vulnerability in the robustness of the models involved.
We hope to drive the active defense of related language models and hope that this attack method is not abused.
We call more attention to the security issues in the information processing of Chinese minority languages.

\begin{acks}
We would like to express our sincere gratitude to the following funding sources for their support of this article: the ``New Generation Artificial Intelligence'' Major Project of Science and Technology Innovation 2030 (No. 2022ZD0116100), the Key Project of Natural Science Foundation of Tibet Autonomous Region ``Research on Adversarial Attack and Defense for Tibetan Natural Language Processing Models'' in 2024, and the Graduate ``High-level Talents Cultivation Program'' Project of Tibet University (No. 2021-GSP-S125).
\end{acks}

\bibliographystyle{ACM-Reference-Format}
\bibliography{socialnlp2024}


\end{document}